\documentclass[journal,onecolumn]{IEEEtran}

\usepackage{cite}
\usepackage{amssymb}
\usepackage{amsgen}
\usepackage{amsfonts}
\usepackage{amsbsy}
\usepackage{amsmath}
\usepackage{amsthm}
\usepackage{color}
\usepackage{subfigure}
\usepackage{tikz} 
\usepackage{graphicx}				
\usepackage{amsmath}
\usepackage{algorithm}
\usepackage{algpseudocode}
\usetikzlibrary{calc}				
\usepackage{amssymb}
\usepackage{graphicx}

\newcommand{\hide}[1]{}

\usepackage{amsfonts}
\usepackage{times}
\usepackage{latexsym}
\usepackage{amssymb}
\usepackage{amsmath}
\usepackage{verbatim}
\newtheorem{theorem}{Theorem}
\newtheorem{corollary}[theorem]{Corollary}
\newtheorem{rem}{Remark}
\newtheorem{lemma}[theorem]{Lemma}
\newtheorem{prop}{Proposition}

\def\bb0{{\mathbb{0}}}

\def\bb{{\mathbf{b}}}

\def\bu{{\mathbf{u}}}

\def\bx{{\mathbf{x}}}

\def\b0{{\mathbf{0}}}

\def\b1{{\mathbf{1}}}

\def\bbR{{\mathbb{R}}}

\def\cB{\mathcal{B}}

\def\cL{\mathcal{L}}

\def\sfL{\mathsf{L}}

\def\sfc{{\mathsf{c}}}

\def\sf0{{\mathsf{0}}}

\def\nn{\nonumber}

\begin{document}

\title{On Dynamic Regret and Constraint Violations in Constrained Online Convex Optimization}
\author{\IEEEauthorblockN{Rahul Vaze}\\
\IEEEauthorblockA{\textit{School of Technology and Computer Science} \\
\textit{Tata Institute of Fundamental Research} \\
\textit{Mumbai, India, rahul.vaze@gmail.com}}
}

\maketitle
\begin{abstract} A constrained version of the online convex optimization (OCO) problem is considered. 
With slotted time, for each slot, first an action is chosen. Subsequently the loss function and the constraint violation penalty evaluated at the chosen action point is revealed. For each slot, both the loss function as well as the function defining the constraint set is assumed to be smooth and strongly convex. In addition, once an action is chosen, local information about a feasible set within a small neighborhood of the current action is also revealed. An algorithm is allowed to compute at most one gradient at its point of choice given the described feedback to choose the next action.  The goal of an algorithm is to simultaneously minimize the dynamic regret (loss incurred compared to the oracle's loss) and the constraint violation penalty (penalty accrued compared to the oracle's penalty). We propose an algorithm that follows projected gradient descent over a suitably chosen set around the current action. We show that both the dynamic regret and the constraint violation is order-wise bounded by 
the {\it path-length}, the sum of the distances between the  consecutive optimal actions. Moreover, we show that the derived bounds are the best possible.

\end{abstract}

\section{Introduction}
Online convex optimization (OCO) has been a very attractive research problem for the last two decades, 
because of its versatility in modelling rich optimization problems. With OCO, at each time $t$, an 
online algorithm selects an action $a_t$, after which the loss incurred $f_t(a_t)$ is revealed. Knowing all $f_t$'s, $1\le t \le T$ ahead of time, an optimal offline algorithm chooses action $x^\star = \arg \min_{x} \sum_{t=1}^T f_t(x)$, and the {\it static} regret 
of an online algorithm is defined as 
$R_s =  \max_{f_t, t=1, \dots, T}\sum_{t=1}^T f_t(a_t) - \sum_{t=1}^T f_t(x^\star),$ i.e., an adversary can choose the functions $f_t$. The name static comes from the 
fact that the optimal offline algorithm is constrained to use a single action. 

Large body of work is known for  static regret \cite{srebro2010smoothness, bartlett2009adaptive, garber2020online, hazan2019introduction,shalev2011online}, where if the functions $f_t$ are convex, the optimal regret is $\Theta(\sqrt{T})$, while if functions $f_t$ are strongly convex
then algorithms with regret at most $O(\log T)$ are known. When functions $f_t$ are chosen 
by an adversary, but the arrival order is randomized, algorithms with better guarantees are also
known \cite{garber2020online}.

Natural generalization of the static regret is the dynamic regret \cite{zinkevich2003online,hall2013dynamical}, where the regret for a comparator sequence $\bu = (u_1, \dots, u_T)$ is defined as 
\vspace{-0.085in}
\begin{equation}\label{defndynamicregret}
R_d(\bu) = \max_{f_t, t=1, \dots, T}\sum_{t=1}^T f_t(a_t) - \sum_{t=1}^T f_t(u_t). 
\end{equation}
For this general dynamic regret definition, sub-linear (in $T$) regret is not always possible, unless some structure is enforced on the comparator sequence. 
For example, for a sequence $\bu$, defining $V_{\bu} =\sum_{t=2}^T ||u_t - u_{t-1}||$,
the online gradient descent (OGD) algorithm was shown to achieve dynamic regret $R_d(\bu) = O(\sqrt{T}(1+V_\bu))$ \cite{zinkevich2003online}, which has been improved to 
$O(\sqrt{T(1+V_\bu)})$ in \cite{zhang2018adaptive}, matching the lower bound $\Omega(\sqrt{T(1+V_\bu)})$ \cite{zhang2018adaptive}. 

A special case of \eqref{defndynamicregret} that is popularly studied \cite{besbes2015non, jadbabaie2015online, yang2016tracking, cdcstrictconvex, zhang2017improved,zhao2020bandit, zhang2020minimizing, zhao2021improved} is by restricting $\bu = \bx^\star =  (x_1^\star, \dots, x_T^\star)$, where $x_t^\star = \arg \min_{x}  f_t(x)$, the sequence of local optimizers. Moreover, in this case, once the action $a_t$ is chosen, the only information available is $\nabla f_t(a_t)$.
For this case, the best known bound on the dynamic regret has been shown to be $O(V_{\bx^\star})$  \cite{cdcstrictconvex} using the online projected gradient descent (OPGD) algorithm, when functions $f_t$ are strongly convex, and smooth. 
Under the special case that the minimizers $x_t^\star$ lie in the interior of the feasible set, $O(V_{\bx^\star})$ regret can be 
achieved when functions $f_t$ are just convex, and smooth \cite{yang2016tracking}. 
For strongly convex and smooth functions, defining $S_{\bx^\star} =\sum_{t=2}^T ||x_t^\star - x_{t-1}^\star||^2$, \cite{zhang2017improved} showed that $O(\min\{V_{\bx^\star}, S_{\bx^\star}\})$ is also 
achievable as long as at each time $t$, gradient of $f_t$ at as many points required is available. Most recently, \cite{zhao2021improved} improved this guarantee to 
$O(\min\{V_{\bx^\star}, S_{\bx^\star}, FV\})$, where 
\vspace{-.1in}
\begin{equation}\label{metricvf}
FV =  \sum_{t=2}^T\sup_{x \in \chi} ||f_t(x) - f_{t-1}(x)||
\end{equation} is the maximum function variation over the feasible set $\chi$ in successive time slots.

In this paper, we focus on the constrained version of the OCO that has been considered more recently in \cite{chen2018bandit, cao2018online, longbo}, where at each time $t$, the objective is to minimize the loss function $f_t$ subject to a constraint $g_t(x)\le 0$. For applications of constrained OCO, we refer to prior work  \cite{chen2018bandit, cao2018online, longbo}. Similar to the unconstrained OCO, even with constraints, the typical information structure 
is that information about $f_t, g_t$ is revealed after the action $a_t$ is chosen. 
The optimizer at time $t$ is $x_t^\star = \arg \min_{x\in \chi:g_t(x) \le 0} f_t(x)$, and the objective is to choose actions to minimize the dynamic regret \eqref{defndynamicregret} with $\bu = \bx^\star$, where the constraint is already absorbed in the definition of $x_t^\star$, and the maximization in \eqref{defndynamicregret} is with respect to both $f_t$ and $g_t$. In addition to the dynamic regret, in this constrained OCO, an additional metric of interest is the constraint violation penalty, which can be defined in many different ways. For example, 
$$P_g(\bx^\star) =  \max_{f_t, g_t t=1, \dots, T}\sum_{t=1}^T ||g_t(x_t^\star)  - g_t(a_t)||,$$ which measures the gap between the function $g_t$ evaluated at the optimal point and the chosen action or $$P_g'(\bx^\star) = \max_{f_t, g_t t=1, \dots, T}\sum_{t=1}^T g_t(a_t),$$ which just counts the overall constraint violation.
We use $P_g(\bx^\star)$ rather than $P_g'(\bx^\star)$ since it is a stronger measure as $P_g(\bx^\star)\ge P_g'(\bx^\star)$ on account of $g_t(x_t^\star)\le 0$.

In prior work, starting from \cite{chen2018bandit}, where functions $f_t, g_t$ are assumed to be convex, Lipschitz and smooth, an algorithm has been proposed that achieves $R_d(\bx^\star) \le O(V_{\bx^\star}\sqrt{T})$ while $P_g'(\bx^\star) = O(T^{1/2})$, which was improved in  \cite{cao2018online}, to get $R_d(\bx^\star) \le O(\sqrt{TV_{\bx^\star}})$ while $P_g'(\bx^\star) = O(V_{\bx^\star}^{1/4}T^{3/4})$, and most recently in  \cite{longbo}, an algorithm based on the drift plus penalty method has 
regret $R_d(\bx^\star) \le O(\max\{\sqrt{TV_{\bx^\star}}, V_g\})$ while $P_g'(\bx^\star) = O(\sqrt{T},V_g)$, or $R_d(\bx^\star) \le O(\sqrt{TV_{\bx^\star}})$ while $P_g'(\bx^\star) = O(T^{3/4},V_g)$, where $V_g$ is as defined in \eqref{metricvf} with $f=g$.

However, notably \cite{longbo} considers the full information setting, where once $a_t$ is chosen, full functions $f_t$ and $g_t$ are revealed, and hence $x_t^\star$ can be computed. Clearly, obtaining this information is highly imposing.
Moreover, \cite{longbo} also needs to know the diameter $D$ of the feasible set. In comparison, the result of \cite{cao2018online} requires the knowledge of $V_{\bx^\star}$ instead of individual $x_t^\star$, which is relatively less demanding, however, still 
very difficult to obtain in practice, as well as the knowledge of $T$ and $D$. 

In this paper, we consider an alternate information structure that is less imposing than considered in \cite{longbo, cao2018online}. The full feasible set at time $t$ is $\chi_t = \{x\in \chi : g_t(x) < 0\}$. We assume that once the current action $a_t$ is chosen, for a fixed constant $ \textsf{dist} > 0 $ that is independent of $T$, a subset of $\chi_t$, set $\chi_t(a_t) = \{x: g_t(x) \le 0\} \cap \cB(a_t, \textsf{dist})$ is made available, where $\cB(x,r)$ is a ball with radius $r$ centered at $x$. Set $\chi_t(a_t)$ captures the feasible set in the neighborhood of the current action. With full information, e.g., in \cite{longbo}, $\textsf{dist} = \infty$. We will show that our results hold for any $\textsf{dist}>0$.


With this new information structure, we consider the problem of simultaneously minimizing the dynamic regret and constraint violation penalty when $f_t,g_t$ are strongly convex, Lipschitz and smooth. Generalizing the results when $f_t,g_t$ are only convex, is part of ongoing work.

Towards this end, we propose an algorithm that uses the projected gradient descent (PGD) algorithm \cite{cdcstrictconvex} as a black box, and depending on the chosen action $a_t$ being feasible 
$g_t(a_t) <0$, on the boundary $g_t(a_t) =0$ or infeasible $g_t(a_t) > 0$, executes PGD over a suitably chosen subset that may or may not be contained in the feasible region of $g_t$. The main concept that the algorithm relies on is the property of the PGD algorithm \cite{cdcstrictconvex} when 
executed over a convex set $I$ and starting point $a_t$,  is that the next action $a_{t+1}$ satisfies 
\begin{equation}\label{eq:contraction}
||x_I^\star - a_{t+1}|| \le \sfc ||x_I^\star- a_{t}||,
\end{equation}
 for a constant $\sfc<1$, where  $x_I^\star = \min_{x\in I} f(x)$  when $f$ is  strongly convex and smooth.

 If the whole feasible region $\chi_t = \{x\in \chi : g_t(x) < 0\}$ was known, then using $I= \chi_t$, \eqref{eq:contraction} will imply that the algorithm is  making `quick' progress 
 towards the optimal point $x_t^\star$. Unfortunately only local information about the feasible region $\chi_t$ is known. In particular, only $\chi_t(a_t) = \chi_t \cap \cB(a_t, \textsf{dist})$ is available for a constant $\textsf{dist}$. Thus, we proceed in two steps. We identify a small 
 region $I_t$ at time $t$ around $a_t$ that is contained in $\chi_t$ and use \eqref{eq:contraction} to claim that we are making progress towards the optimal point in this subset $I_t$ (which could be far away from the global optimal). Next, exploiting the strong convexity and the smoothness of the functions, we extend the same claim to the optimal point $x_t^\star$ which need not be in $I_t$. 
 
 Since we have only local information about $g_t$ around $a_t$,
 it can happen that the size of $I_t$ is arbitrarily small or $I_t$ is empty in case $g_t(a_t) >0$ (current choice is infeasible). For both these cases, we 
 show that the algorithm makes progress of a finite distance towards the optimal point $x_t^\star$ in $\chi_t$, and establish a relation similar to \eqref{eq:contraction}. Once we have \eqref{eq:contraction}, a simple application of the triangle inequality and the Lipschitz condition, implies the result.
 
 Our contributions.
\begin{itemize}
\item We show that under the defined information structure, the proposed algorithm simultaneously achieves $R_d(\bx^\star) \le O(V_{\bx^\star})$ and $P_g'(\bx^\star) \le P_g(\bx^\star) \le O(V_{\bx^\star})$ for any $\textsf{dist}>0$. Importantly, no information about $x_t^\star, V_{\bx^\star}, T$ or $D$ is needed.
\item 
As a function of information variable $\textsf{dist}>0$, both $R_d(\bx^\star)$ and $P_g(\bx^\star)$ scale inverse polynomially, which is natural to expect since for  any algorithm as information availability is decreased, (smaller value of $\textsf{dist}$), the regret should worsen. We do not know at this point if the algorithm achieves the best scaling in terms of $\textsf{dist}$.
\item In Remark \ref{rem:bestpossible}, we also argue  that our result is the best one can hope for, given the minimal information structure.
\end{itemize}

Notation: For the rest of the paper, we follow the notation described as follows. For a set $I \in \bbR^n$, its interior is defined as $\text{int}(I)$, while its boundary as $\text{boundary}(I)$. 
 $\cB(x,r)$ is the ball of radius $r$ centered at $x$. For a discrete set of points $S$, $\text{convex hull}(x\in S)$ represents the convex hull of points $x\in S$.
 $\text{Proj}(x, S)$ is the projection of point $x$ on set $S$, i.e. $\text{Proj}(x, S) = \arg \min_{y\in S} ||x-y||$.

\section{System Model}
Time is slotted with total time horizon $T$, and time slots are indexed as $t=1,\dots, T$. Let $\chi \subset \bbR^n$ be a compact and convex set. For each $t$, two functions $f_t$ and $g_t$ are of interest, that are defined over $\chi$.
The feasible set at time $t$ is defined as $\chi_t = \{x \in \chi: g_t(x) \le 0\}$. Let the optimizer for $f_t$ over the constraint set $g_t(x) \le 0$ be $x_t^\star$, i.e., $x_t^\star = \arg \min_{\{x \in \chi_t\}} f_t(x)$.  

We make the following standard assumptions about $f_t$ and $g_t$.
Functions $f_t$ and $g_t$ are assumed to be Lipschitz with Lipschitz constants $\cL_f$ and $\cL_g$, respectively.
Moreover, functions $f_t$ and $g_t$ are assumed to be smooth, i.e., the gradients $\nabla f_t$ and $\nabla g_t$ are assumed to be Lipschitz with Lipschitz constants $L_f$ and $L_g$, respectively. Moreover, for all $1\le t\le T, \sup_{ x\in \chi} ||\nabla f_t(x)|| \le G$ and $\sup_{ x\in \chi}||\nabla g_t(x)|| \le G$. \footnote{For notational simplicity we are assuming the same constant $G$, which can be generalized without any change in following analysis.} Compared to prior work  \cite{longbo, cao2018online} that assume that $f_t$ and $g_t$ are convex,  we assume that $f_t$ and $g_t$ are {\it strongly} convex with strong convexity parameters $\nu_{f}, \nu_{g}$, respectively. \footnote{ We are assuming that  all $f_t$'s and $g_t$'s have the same smoothness parameter $L_f$ and $L_g$ only for notational simplicity. All results will go through with different parameters as well.}

At each time $t$, an action $a_t$ is chosen by an algorithm, for which the cost is  
$f_t(a_t)$. The goal of the algorithm to choose $a_t$ such that the cost $f_t(a_t)$ is as small as possible while making sure that $a_t\in \chi_t$. However, the information available with the algorithm to choose $a_t$ is limited and described as follows.

Information structure:
Similar to \cite{chen2018bandit, cao2018online, longbo}, once the action $a_t \in \bbR^n$ is chosen at time $t$, $g_t(a_t)$ is revealed. Moreover, the algorithm can also access $\nabla f_t(x),  \nabla g_t(x)$ for at most one point  $x$ of its choice. As described in the Introduction, additionally, in this paper, we assume that, set $\chi_t(a_t) = \chi_t \cap \cB(a_t, \textsf{dist})$ is also revealed at time $t$ for a fixed constant $ \textsf{dist} > 0 $, after $a_t$ has been chosen. 
Note that $\textsf{dist}$ can be arbitrarily 
small but is a constant that is fixed throughout the time horizon and does not depend on $t$ or $T$. Compared to prior work, \cite{chen2018bandit, cao2018online, longbo}, acquiring this  information is less imposing and does not involve finding any $x_t^\star$. The set $\chi_t(a_t)$ maps the local behaviour of $g_t$ in a very small neighborhood of $a_t$. Note that convexity implies that $\chi_t(a_t)$ is convex for any $a_t$.

\begin{rem}
For the considered problem to be meaningful, once $a_t$ is chosen, $g_t(a_t)$ has to be revealed, as already assumed  in prior work \cite{chen2018bandit, cao2018online, longbo}. In this work, in addition, we are assuming that $\chi_t(a_t)$ is also known which in turn requires that $g_t(x)$ for $x\in B(a_t, \textsf{dist})$ is also known. When $\textsf{dist} \rightarrow 0$, this new information is equivalent to just acquiring $g_t(a_t)$. Since $\textsf{dist}$ is allowed to be any arbitrarily small constant, the extra information assumed is very minimal and can be obtained similar to obtaining $g_t(a_t)$ (necessary), and can be done efficiently by exploiting the convexity of $g_t$.
\end{rem}

The performance metric for an online algorithm that chooses actions $a_t, t=1, \dots, T$ is defined as the dynamic regret  
$$R_d(\bx^\star) = \max_{f_t,g_t}\sum_{t=1}^T ||f_t(x_t^\star)  - f_t(a_t)||,$$ 
and penalty for constraint violation as 
$$P_g(\bx^\star) = \max_{f_t,g_t} \sum_{t=1}^T ||g_t(x_t^\star)  - g_t(a_t)||,$$
where $a_t$'s are the causal actions of the algorithm that can depend on the information acquired till time slot $t-1$. Moreover, $f_t,g_t$ can be chosen by an adversary (can be adaptive, i.e., depend on previous actions $a_\tau, \tau\le t-1$) and are not required to follow any structure, other than what has been described earlier. 

Note that $P_g$ is stronger than the penalty considered in earlier work \cite{longbo} that is defined as 
$P_g'(\bx^\star) = \max_{f_t,g_t}\sum_{t=1}^T g_t(a_t),$ in two aspects. $P_g'(\bx^\star)$ can be negative, while $P_g(\bx^\star)$ is always positive, and $P_g(\bx^\star)\ge P_g'(\bx^\star)$ since $g_t(x_t^\star)$ can be 
negative.

\section{Algorithm}
We present the proposed algorithm as a pseudo code in Algorithm \ref{alg}, and describe it as follows. 
Let at time $t$, $$x_t^\star = \arg \min_{x\in \chi_t} f_t(x).$$
Let the  action chosen at time $t$ be $a_t$. We want to choose $a_{t+1}$ in such a way that 
\begin{equation}\label{eq:wanted}
||x_t^\star-a_{t+1}||^2 < c ||x_t^\star-a_{t}||^2,
\end{equation}
for some constant $0< c< 1$ that does not depend on $t$.  Recall that  while choosing $a_{t+1}$, no information about $f_{t+1}, g_{t+1}$ is available.  Thus, relation \eqref{eq:wanted} is useful in the sense that it ensures that $a_{t+1}$ is closer to $x_{t}^\star$ compared to $a_t$, in hope that if $x_t^\star$ and $x_{t+1}^\star$ are close, then $a_{t+1}$ will be close to $x_{t+1}^\star$ as well.

The main idea of the algorithm is to accomplish this goal (showing that \eqref{eq:wanted} holds) depending on three possible cases, namely : 
i) $g_t(a_t) < 0$, i.e. $a_t$ is strictly feasible for $g_t$, ii) $g_t(a_t) =0$, i.e. $a_t$ is on the boundary of the feasible region for $g_t$, and finally, iii) $g_t(a_t) > 0$, i.e. $a_t$ is strictly infeasible for $g_t$. Just to be clear, all the described actions in the following are taken after $a_t$ is chosen and the information has been 
revealed about $\nabla f_t(x), g_t(a_t)$, $\nabla g_t(x)$ and  $\chi_t(a_t)$ for some one point $x$.

In case i) $g_t(a_t) < 0$, and we know that $a_t$ is strictly feasible and potentially there is room to move to a point closer to $x_t^\star$, the optimizer of $f_t$. Using the Lipschitz property of $g_t$'s, this implies that each point in the ball $\cB(a_t, ||\frac{g_t(a_t)}{2\cL_g}||)$ is also feasible. Thus, Algorithm \ref{alg} chooses the set $\cB(a_t, ||\frac{g_t(a_t)}{2\cL_g}||)$ as the feasible region to execute the PGD. 

In case, the radius $||\frac{g_t(a_t)}{2\cL_g}||$ of the identified feasible region is smaller than the fixed constant $\textsf{dist}$, then using the extra local information 
$\chi_t(a_t)$ as described earlier, the feasible region is chosen as $\chi_t(a_t)$.
A local gradient descent algorithm over the chosen feasible region using subroutine \textsc{Optimize} Algorithm \ref{alg:opt} (online gradient descent) is 
used to find the next action $a_{t+1}$.
 
In case ii) $g_t(a_t) =0$, $a_t$ is on the boundary of the feasible region. In this case, we use the local information about $g_t$ around $a_t$ and choose $\chi_t(a_t)$ as the feasible region. Next,  a
local gradient descent is executed using subroutine \textsc{Optimize} Algorithm \ref{alg:opt} in the identified feasible region to find the 
next action.

Finally in case iii) $g_t(a_t) >0$ $a_t$ is strictly infeasible. Since the current choice of $a_t$ is infeasible for $g_t$, and $g_t(x_t^\star) \le 0$, it is sufficient to move towards the region for which $g_t(x) \le 0$ to ensure \eqref{eq:wanted} while staying infeasible. In fact, if we `blindly' move into the feasible region, we cannot guarantee that  $a_{t+1}$ is closer to $x_t^{\star}$ than 
$a_t$, for example if $g_t(x_t^\star) = 0$. However, using the Lipschitz condition, we know that each point in the ball $\cB(a_t, ||\frac{g_t(a_t)}{2\cL_g}||)$ is infeasible given that $a_t$ is infeasible. Thus, in this case as long as $||\frac{g_t(a_t)}{2\cL_g}||\ge ||\nabla g_t(a_t) \frac{1}{L_g}||$ 
we move a distance of $||\nabla g_t(a_t) \frac{1}{L_g}||$ from $a_t$ in the direction of negative 
gradient of $g_t$ at $a_t$. Thus the new point $a_{t+1}$ is still infeasible, but as we show in Lemma \ref{lem:case3}, $a_{t+1}$ is closer 
to $x_t^\star$ than $a_t$. In case  $||\frac{g_t(a_t)}{\cL_g}|| < ||\nabla g_t(a_t) \frac{1}{L_g}||$, the algorithm 
finds a feasible region similar to case i) using the local information $\chi_t(a_t) $  and follow a 
local gradient descent in this feasible region using subroutine \textsc{Optimize} Algorithm \ref{alg:opt} to find the next action. 
In case, $\chi_t(a_t)$ turns out to be an empty set, we proceed similar to the case when  $||\frac{g_t(a_t)}{\cL_g}||\ge ||\nabla g_t(a_t) \frac{1}{L_g}||$, since the whole of $\cB(a_t, \textsf{dist})$ is infeasible.

%
 \begin{algorithm}
\caption{Algorithm}\label{alg}
\begin{algorithmic}[1]
\State Input $L_f, L_g, \cL_g, \textsf{dist} > 0$, feasible set $\chi \subset \bbR^n$
\State Initialize $t=0$
\State Choose action $a_1$ arbitrarily belonging to $\chi$

\While{$t\le T-1$}

\State $t=t+1$

\If{$g_t(a_t) < 0$} \%previous action $a_t$ was strictly feasible for $g_t$
\State $\delta_t = ||\frac{g_t(a_t)}{2\cL_g}||$
\If  {$\delta_t \ge \textsf{dist}$ }
		\State  $I_t = \cB(a_t, \delta_t)$
		\State $a_{t+1}=\textsc{Optimize}(f_t, I_t,\frac{1}{2L_f},a_t)$  
		\Else 
		\State Find the feasible region $\chi_t(a_t) = \chi_t \cap \cB(a_t, \textsf{dist})$  \State 
$I_t= \chi_t(a_t)$
\State $a_{t+1}=\textsc{Optimize}(f_t, I_t,\frac{1}{2L_f},a_t)$  
	\EndIf
	
		\ElsIf{$g_t(a_t) =0$} \%$a_t$ is on the boundary of the feasible region
		
		\State Find the feasible region $\chi_t(a_t) = \chi_t \cap \cB(a_t, \textsf{dist})$  \State 
$I_t= \chi_t(a_t)$
\State $a_{t+1}=\textsc{Optimize}(f_t, I_t,\frac{1}{2L_f},a_t)$

		\ElsIf{$g_t(a_t) > 0$} \%$a_t$ is infeasible
		\If {$ \delta_t \ge ||\nabla g_t(a_t) \frac{1}{L_g}||$ }
		\State  $a_{t+1} =   a_t + \alpha ({\hat a}_{t} - a_t)$, where  - 
		\State ${\hat a}_{t} = a_t - \nabla g_t(a_t) \frac{1}{L_g}$, 
		
				\Else \quad \State Find the feasible region $\chi_t(a_t) = \chi_t \cap \cB(a_t, \textsf{dist})$ 
		\If {$\chi_t(a_t) \neq \emptyset$} 
		\State $I_t= \chi_t(a_t)$
		\State $a_t' = \text{Proj}(a_t, I_t)$
		 
\State $a_{t+1}=\textsc{Optimize}(f_t, I_t,\frac{1}{2L_f},a_t')$ 
\Else 
\State  $a_{t+1} =   a_t + \alpha ({\hat a}_{t} - a_t)$, where  -
\State ${\hat a}_{t} = a_t- \nabla g_t(a_t)~\frac{\textsf{dist}}{||\nabla g_t(a_t)||}$
\EndIf
		\EndIf
\EndIf
\EndWhile
\end{algorithmic}
\end{algorithm}

\begin{theorem}\label{thm1}
When both $f_t, g_t$ are strongly convex, Lipschitz, and smooth  for all $t\le T$  and $1\le t\le T, \sup_{ x\in \chi}\nabla f_t(x) \le G$ and $\sup_{ x\in \chi}\nabla g_t(x) \le G$, with information structure as defined, for Algorithm \ref{alg}
$$||x_t^\star-a_{t+1}|| < c  ||x_t^\star-a_{t}||,$$
for some constant $0< c< 1$ that does not depend on $t$. In particular, $$c = \max\left\{c_2, c_3, c_4, c_5\right\}< 1,$$ where $c_2 = \left(1-\frac{\alpha \nu_f}{2L_f}\right)^{1/2}, c_3 =\frac{D+ \alpha \textsf{dist}}{D+\textsf{dist}}$ and 
$c_4 = (1-\alpha \nu_g/L_g)^{1/2}, c_5 =  \left(1-\alpha \frac{\nu_g}{\max\{G/\textsf{dist}, L_g\}}\right)^{1/2}$, and 
$0< \alpha < 1$ is a constant to be chosen by subroutine \textsc{Optimize}, $D$ 
is the diameter of the feasible region and $0< \textsf{dist}$ is a constant to be chosen by Algorithm \ref{alg}.
Note that $\nu_{g} \le L_g$ and $\nu_f \le L_f$ always, thus $0 < c < 1$.
\end{theorem}
\begin{algorithm}
\caption{\textsc{Optimize}}\label{alg:opt}
\begin{algorithmic}[1]
\State Input$(h,I,\mu, x_t)$
\State  Constant $0< \alpha < 1$,
$x_{t+1} =  x_t + \alpha ({\hat x}_t - x_t)$,
where 
$${\hat x}_t = \text{Proj}( x_t - \frac{1}{\mu}  \nabla h(x_t), I).$$
\end{algorithmic}
\end{algorithm}

Using Theorem \ref{thm1}, we get the main result of the paper as follows.

 \begin{theorem}\label{thm2}When both $f_t, g_t$ are strongly convex, Lipschitz, and smooth ($\nabla f_t, \nabla g_t$ are Lipschitz) for all $t\le T$ and  $1\le t\le T, \sup_{ x\in \chi}\nabla f_t(x) \le G$ and $\sup_{ x\in \chi}\nabla g_t(x) \le G$, with information structure as defined,
 with Algorithm \ref{alg}, simultaneously, 
$$R_d(\bx^\star) = O(V_{\bx^\star}), \text{and}, \ P_g(\bx^\star) = O(V_{\bx^\star}).$$
\end{theorem}
\begin{rem}
Both the regret and constraint violation penalty bounds derived in Theorem \ref{thm2} are inverse polynomially  proportional to the chosen constant $\textsf{dist}$. In particular, they grow as $\frac{1}{1-c_5}$ where $c_5 =  \left(1-\alpha \frac{\nu_g}{\max\{G/\textsf{dist}, L_g\}}\right)^{1/2}$. 
It is natural to expect that regret grows with decreasing $\textsf{dist}$ since for  any algorithm as information availability is decreased, (in this case smaller value of $\textsf{dist}$), the regret should worsen. However, $\textsf{dist}$ can be any constant and not necessarily has to be $<<1$, and there is a tradeoff between regret and the amount of available local feasibility information $B(a_t, dist)$.
\end{rem}
\begin{rem}\label{rem:bestpossible}
For the unconstrained OCO, when at each step gradient information is available only at a single point, the best known algorithm when each $f_t$ is smooth, and strongly convex, has $R_d(\bx^\star) \le O(V_{\bx^\star})$ \cite{cdcstrictconvex}.
Note that $V_{\bx^\star}$ in the constrained and the unconstrained 
OCO problem are different, therefore directly we cannot compare our result with that of \cite{cdcstrictconvex}. However, since functions $g_t$ and $f_t$ are allowed to be arbitrary with the constrained OCO, $g_t=f_t$ for each $t$ is a valid choice for $g_t$ and $f_t$. With $g_t=f_t$, the constrained OCO collapses to the unconstrained OCO, for which the best known result on regret is $O(V_{\bx^\star})$, making the derived result (which also needs gradient availability at only one point) the best possible.

\end{rem}

\begin{proof} [Proof of Theorem \ref{thm2}]
Using the triangle inequality, we get that $\sum_{t=1}^T ||x_t^{\star} - a_{t}||$
\begin{align}\nn
 & \le   || x_1^{\star} -a_1 || + \sum_{t=2}^T  || x_{t-1}^{\star} - a_t || +  \sum_{t=2}^T  ||x_t^{\star} - x_{t-1}^\star||, \\\nn
& \stackrel{(a)} \le ||x_1^{\star} - a_{1}|| + c \sum_{t=2}^T  ||x_{t-1}^{\star} - a_{t-1}|| +  \sum_{t=2}^T  ||x_t^{\star} - x_{t-1}^\star||, \\ \nn
& \stackrel{(b)} \le ||x_1^{\star} - a_{1}|| - c||x_T^{\star} - a_{T}|| + c \sum_{t=1}^T  ||x_{t}^{\star} - a_{t}|| \\
& \quad \quad +  \sum_{t=2}^T  ||x_t^{\star} - x_{t-1}^\star||, \label{dec4}
\end{align}
where $(a)$ is obtained by using Theorem \ref{thm1}, while to obtain $(b)$ we added and subtracted $c||a_T- x_T^\star||$ and rearranged terms.

Regrouping terms in \eqref{dec4}, we get $\sum_{t=1}^T ||x_t^{\star} - a_{t}||$
\begin{equation}\label{dec5}
  \le \frac{ ||x_1^{\star} - a_{1}|| - c ||x_T^{\star} - a_{T}||}{1-c} 
+ \frac{1}{1-c} \sum_{t=2}^T ||x_t^{\star} - x_{t-1}^{\star}||.
\end{equation}

Thus, using the Lipschitz property of $f_t$ and $g_t$, \eqref{dec5} implies that $R_d(\bx^\star) =\sum_{t=1}^T ||f_t(x_t^{\star}) - f_t(a_{t})||$
$$ \le  \frac{L_f}{1-c} \sum_{t=2}^T ||x_t^{\star} - x_{t-1}^{\star}|| + \frac{D}{1-c}= O(V_{\bx^\star})$$ 
and $P_g(\bx^\star) =\sum_{t=1}^T ||g_t(x_t^{\star}) - g_t(a_{t})||$
$$  \le  \frac{L_g}{1-c} \sum_{t=2}^T ||x_t^{\star} - x_{t-1}^{\star}||+ \frac{D}{1-c}= O(V_{\bx^\star}),$$
where  $V_{\bx^\star} = \sum_{t=2}^T ||x_t^{\star} - x_{t-1}^{\star}||$, the accumulated variation of the per-step minimizers.
\end{proof}

Next, we first briefly discuss the basic difference between the proposed algorithm and the relevant prior work.
In \cite{cao2018online}, a primal dual algorithm has been proposed using the  Lagrangian 
$$\sfL(x,\lambda) = f_t(x) + \lambda^Tg_t(x)+\frac{\eta}{||\lambda||^2},$$ where $a_t$ is updated using
the gradient descent over the Lagrangian to move towards the optimizer of $f_t$ with penalty function $\lambda$ as 
$$a_{t+1} = a_t - \eta \nabla_x \sfL(x,\lambda),$$
while gradient ascent is used to increase the penalty in case of constraint violation as $\lambda_{t+1} = \lambda_t - \eta \nabla_\lambda \sfL(x,\lambda)$.

Similarly, in \cite{longbo}, a primal dual algorithm is proposed where the increase in $\lambda$ is derived by minimizing the expected `drift' of the constraint 
violation. In particular, it is given by  
$$\lambda_{t+1} = \max\{\lambda_{t} + \eta_{t} g_t(a_t), -  \eta_{t} g_t(a_t)\},$$ while 
$$a_{t+1} =  \nabla f_t^T(a_t)(x-a_t) +\mu_t||x-a_t||^2 +  [\lambda_t + \eta_{t}g_t(a_t)]^T\eta_{t}g_t(a_t).$$

Both these algorithms \cite{cao2018online, longbo} are long-term in the sense that they want to remain close to $x_t^\star$ while minimizing the constraint violation penalty $P_g'(\bx^\star)$ 
in the long term, i.e., they nudge the updates `slowly' in the direction of constraint satisfaction to avoid large accumulated constraint violation penalty. In contrast, the proposed algorithm in this paper is local, and is trying to go close to the optimal point in every single step as shown in Lemma \ref{lem:case1}, 
\ref{lem:case2} and \ref{lem:case3}. Thus, conceptually our algorithm is entirely different than \cite{cao2018online, longbo}. 

In terms of restrictions, over and above \cite{cao2018online, longbo}, we assume that $f_t$ and $g_t$ are strongly convex, however in terms of information, we require far less. In particular, at time $t$, after $a_t$ has been chosen, Algorithm \ref{alg} requires only $g_t(a_t)$, $\chi_t(a_t)$ and $\nabla f_t(x), \nabla g_t(x)$, at $x=a_t$ or some $x\in \chi_t(a_t)$. In contrast, \cite{longbo} assumes that once $a_t$ is chosen, full $f_t, g_t$ are revealed, making $x_t^\star$ known. Moreover, it requires the knowledge of the diameter $D$. In \cite{cao2018online}, knowledge of $V_{\bx^\star}, D, T$ is needed over and above $\nabla f_t(a_t), \nabla g_t(a_t), g_t(a_t)$.

In the rest of the paper, we prove Theorem \ref{thm1}, for which we need the following Lemma regarding the subroutine \textsc{Optimize}.
\begin{lemma}\cite{cdcstrictconvex}\label{lem:mother}
If function $h$ is $\nu_h$-strongly convex, and $\nabla h$ is Lipschitz with parameter $L_h$, and $x^{I\star} = \arg \min_{x\in I} h(x)$, then if  parameter $\mu \ge L_h$, the output $x_{t+1}$ from subroutine \textsc{Optimize} satisfies
  \begin{equation}\label{eq:mother}
||x^{I\star} - x_{t+1}|| \le \sfc ||x^{I\star} - x_{t}||,
\end{equation}
for $\sfc = \left(1-\alpha \frac{\nu_h}{\mu}\right)^{1/2}<1$.

\end{lemma} 

\begin{corollary}\label{lem:mothercor}
For subroutine \textsc{Optimize},  let ${\tilde x} \in I$ be such that $h({\tilde x}) < h({\hat x}_{t})$, then  
with parameter $\mu \ge L_h$, the output $x_{t+1}$ from subroutine \textsc{Optimize} satisfies 
\begin{equation}\label{eq:mothercor}
||{\tilde x} - x_{t+1}|| \le \sfc ||{\tilde x} - x_{t}||,
\end{equation}
for $\sfc = \left(1-\alpha \frac{\nu_h}{\mu}\right)^{1/2}<1$ 
as long as function $h$ is $\nu_h$-strongly convex, and $\nabla h$ is Lipschitz with parameter $L_h$.
\end{corollary} 

\begin{proof} The only place where optimality of $x^{I\star}$ is used in the proof of Lemma \ref{lem:mother} in \cite{cdcstrictconvex} is to 
show that $h({\hat x}_t) > h(x^{I\star})$. Thus, the proof goes through as it is, even with this weaker condition that $h({\tilde x}) < h({\hat x}_{t})$. For completeness, the full proof is given in Section \ref{appcor}.
Another way to see the result is that by pruning $I$ to get $I'$ such that ${\tilde x} = \arg \min_{I'} h(x)$, while keeping $h$ a strongly convex function over $I'$. Thus, the result follows directly from Lemma \ref{lem:mother}.
\end{proof} 

For ease of exposition, we break the proof of Theorem \ref{thm1} into three parts corresponding to $g_t(a_t) < 0, g_t(a_t) = 0$, and $g_t(a_t) >0$ in the next three lemmas.  
\begin{lemma}\label{lem:case1}
When both $f_t, g_t$ are strongly convex and smooth  for all $t\le T$ with information structure as defined, with Algorithm \ref{alg}, for the case when $g_t(a_t) < 0$
$$||x_t^\star-a_{t+1}|| < c_1 ||x_t^\star-a_{t}||,$$
where  $0< c_1 = \max\{c_2, c_3\} < 1$  for $c_2 = \left(1-\frac{\alpha \nu_f}{2L_f}\right)^{1/2}, c_3 =\frac{D+ \alpha \textsf{dist}}{D+\textsf{dist}}< 1$ that does not depend on $t$. Since $L_f \ge \nu_f$ (always), $c_2 < 1$.
\end{lemma}
\begin{proof}
Recall that $\delta_t = ||\frac{g_t(a_t)}{2\cL_g}||$.

Case a) $\delta_t \ge \textsf{dist}$. In this case, $I_t= \cB(a_t, \delta_t)$  and $I_t \subseteq \chi_t$ using the Lipschitz condition on $g_t$. 

Subroutine \textsc{Optimize} is executed with set $I_t$ and starting point $a_t$. The output of Subroutine \textsc{Optimize} is 
\begin{equation}\label{updatef2}
a_{t+1} = a_t + \alpha({\hat a}_{t} - a_t),
\end{equation}
$${\hat a}_{t} = \text{Proj}(a_t - \nabla f_t(a_t) \frac{1}{2L_f}, I_t).$$

Subcase a-i) ${\hat a}_{t}  \in \text{convexhull}(a_t, x_t^\star)$  (just the line segment connecting $a_t$ and $x_t^\star$). 
If $x_t^\star \in I_t$, then directly from Lemma \ref{lem:mother}, we get 
\begin{equation}\label{dummy1xx1}
||a_{t+1} - x_t^\star||\le c_2 ||a_{t} - x_t^\star||,
\end{equation}
where $c_2 = \left(1-\alpha \frac{\nu_f}{2L_f}\right)^{1/2}$ as we have chosen  $\mu = 2L_f$.

Otherwise, if $x_t^\star \notin I_t$, then ${\hat a}_{t}\in \text{boundary}(I_t)$ since $f_t$ is strongly convex, ${\hat a}_{t}  \in \text{convexhull}(a_t, x_t^\star)$  and $x_t^\star \notin I_t$. Thus, the distance between ${\hat a}_{t}$ and $a_t$ is at least $\textsf{dist}$ since $\delta_t \ge \textsf{dist}$, and the distance between $a_{t+1}$ and $a_t$ is at least $\alpha \textsf{dist}$, while the distance between $a_{t+1}$ and $x_t^\star$ is at most $D$ (the diameter). Thus, we get that 
\begin{equation}\label{dummy1xx1}
||a_{t+1} - x_t^\star||\le c_3 ||a_{t} - x_t^\star||,
\end{equation}
where $c_3 = \frac{D+ \alpha \textsf{dist}}{D+\textsf{dist}}$.

Subcase a-ii) ${\hat a}_{t}  \notin \text{convexhull}(a_t, x_t^\star)$
 
If $x_t^\star \in I_t$, then directly from Lemma \ref{lem:mother}, we  get that
\begin{equation}\label{dummy1xx12}
||a_{t+1} - x_t^\star||\le c_2 ||a_{t} - x_t^\star||,
\end{equation}
as we have chosen $\mu = 2L_f$.

Thus, consider the case when $x_t^\star \notin I_t$.
Let $I_t' = \text{convex hull}(a_t, {\hat a}_{t}, x_t^\star)$ where $I_t' \subseteq \chi_t$, i.e. full set $I_t' $ is feasible, since $g_t$ is convex. 

Now, consider that if Subroutine \textsc{Optimize} is executed with set $I_t'$ and the same starting point $a_t$, the output of Subroutine \textsc{Optimize} will be the same as  \eqref{updatef2}, since 
$$\text{Proj}(a_t - \nabla f_t(a_t) \frac{1}{2L_f}, I_t) = \text{Proj}(a_t - \nabla f_t(a_t) \frac{1}{2L_f}, I_t'),$$
irrespective of whether $ a_t - \nabla f_t(a_t) \frac{1}{2L_f}$ belongs to $I_t$ or not. However, since $x_t^\star \in I_t'$, we get from Lemma \ref{lem:mother} that
\begin{equation}\label{dummy1xx}
||a_{t+1} - x_t^\star||\le c_2 ||a_{t} - x_t^\star||.
\end{equation}
An illustration of the basic idea of the proof when ${\tilde a}_{t} \notin I_t$ is presented in Fig. \ref{fig:technical}.

\begin{figure}
\begin{centering}
\begin{tikzpicture}[
    mycirc/.style={circle,fill=black,minimum size=0.005cm,scale=0.4}
    ]
\draw[rotate around={-45:(7,2)},red,dashed] (7,2) ellipse (20pt and 40pt);

    \node[mycirc,label=left:{$a_t$}] (s1) at ($(7,2)$) {};
        \node[mycirc,label=above:{${\hat a}_t= \text{Proj}({\tilde a}_t, I_t)$}] (projs1) at ($(7.89,2)$) {};

    \node[mycirc,label=right:{$ {\tilde a}_{t}=a_t - \nabla f_t(a_t) \frac{1}{2L_f} $}] (hats1) at ($(9,2)$) {};
        \node[mycirc,label=left:{$x_t^{I_t \star}$}] (l) at (6,1) {};


\node[text width=3cm] at (7.8,1.25) 
    {$I_t$};
    \node[text width=3cm, blue] at (8.9,1.25) 
    {$I_t'$};

\node[mycirc,label=right:{$x_t^\star$}] (z3) at ($(7.57,-.5)$) {};
    \draw [dashed,blue] (s1) -- (z3);
    \draw [dashed, blue] (z3) -- (7.89,2);
    \draw [dashed, blue] (projs1) -- (s1);
 \draw [->, dashed, blue] (hats1) -- (projs1) ;
\end{tikzpicture}
\caption{Illustration for the proof of  Lemma \ref{lem:case1} case a-ii), where the blue dashed triangle is $I_t' = \text{convex hull}(a_t, {\hat a}_{t}, x_t^\star) \subseteq \chi_t$.}
\label{fig:technical}
\end{centering}
\end{figure}
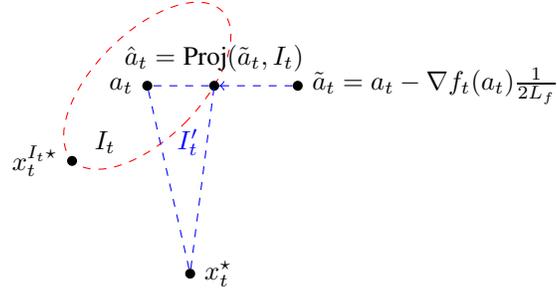



Case b) $ \delta_t  < \textsf{dist}$ Except for the choice of set $I_t$ which is now $\chi_t(a_t)$ everything else is same as in case a). Moreover, since $\chi_t(a_t) \in \chi_t$ is feasible by definition, the same arguments as detailed in  case a) apply, and we either get \eqref{dummy1xx1} or \eqref{dummy1xx}.

The two distinct choices of $I_t$ are essentially made to speed up the algorithm. Always choosing $I_t = \chi_t(a_t)$ is sufficient for analysis.

\end{proof}

At this point it is difficult to appreciate the power of Lemma \ref{lem:case1}. What Lemma \ref{lem:case1} saying is that irrespective of the size (how small) of set $I_t$ chosen by the algorithm, as well as independent of the distance of $x_t^\star$ (however far) from $I_t$, we get a relation \eqref{dummy1xx}, that states that the distance between the optimal point and the updated point $a_{t+1}$ contracts by a fixed amount compared to the original point $a_t$. 
The main tool that we are exploiting to prove Lemma \ref{lem:case1} is both the strong convexity as well as the smoothness (gradient being Lipschitz) of the function $f_t$, and in some measure of $g_t$. To gather more intuition we consider a one-dimensional case in Figs. \ref{fig:intuition1} and \ref{fig:intuition2} to show how strong convexity together with smoothness indicates that contraction of distance from the optimal holds independent of the distance between the present point $a_t$, the updated point $a_{t+1}$, and the optimal point $x_t^\star$. 

In Fig. \ref{fig:intuition1},  for function $f(x)$ which is assumed to be strongly convex and smooth, we consider that the feasible set is $\chi_1 = (-\infty, x_1)$ and $x^\star = x_1$, while in Fig. \ref{fig:intuition2} it is $\chi_2 = (-\infty, x_2)$ and $x^\star = x_2$. Clearly, by construction, $a_{t+1}$ remains the same when  \textsc{Optimize} is executed with starting point $a_t$, input function $h=f$ with $I = \chi_1$ or $\chi_2$ and an identical choice of $\mu$. Thus, from Lemma \ref{lem:mother}, 
we get that 
\begin{equation}\label{ex1}
||a_{t+1} - x_1||\le c ||a_{t} - x_1||.
\end{equation}
as well as 
\begin{equation}\label{ex2}
||a_{t+1} - x_2||\le c||a_{t} - x_2||.
\end{equation}
for the same $c<1$.
Clearly, as $x_2$ is moved sufficiently far away to the right, one does not expect \eqref{ex2} to hold together with \eqref{ex1}. However, since $f$ is both strongly convex and smooth, there is a limit on how far $x_2$ can be compared to $x_1$, before $f$ starts to increase. This is the key reason behind both \eqref{ex1} and \eqref{ex2} to be true. Essentially, when $f$ is both strongly convex and smooth, it is `trapped' between a lower and an upper envelope.

In general, coming back to Lemma \ref{lem:case1}, because of the strong convexity and smoothness, the function $f_t$ cannot continue to decrease beyond a point, and the estimate one gets for the contraction in \eqref{dummy1xx} is an underestimate when $x_t^\star$ is close to $a_t$ and $a_{t+1}$, while becomes tighter as $x_t^\star$ is drawn away from $a_t$ and $a_{t+1}$.


\begin{figure}[t!]
    \centering
        \centering
        \begin{tikzpicture}[scale=0.6]
         \draw[-] (.6,-0.4) node[right] {$a_t$};
          \draw[-] (1.6,-0.4) node[right] {$a_{t+1}$};
        
        \draw[-] (3.6,-0.4) node[right] {$x_1$};
        \draw[-] (-1.5,4) node[right] {$f(x)$};
\draw (4,-0.2) -- (4, 1.62);
\draw (2,-0.2) -- (2, 2.24);
\draw (1,-0.2) -- (1, 3.48);
  \draw[->] (-1, 0) -- (6.2, 0) node[right] {};
  \draw[->] (0, -1) -- (0, 4.2) node[above] {};
  

      \draw[scale=0.5, domain=1:8, smooth, variable=\x, blue] plot ({\x}, { 1/\x)*10+2});  

\end{tikzpicture}
        \caption{$\chi_1 = (-\infty, x_1)$}
  \label{fig:intuition1}
    \end{figure}
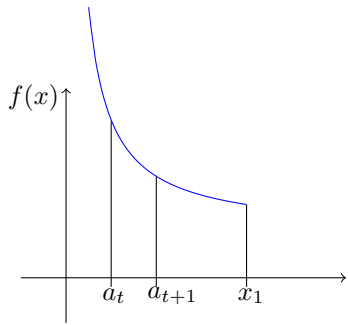
    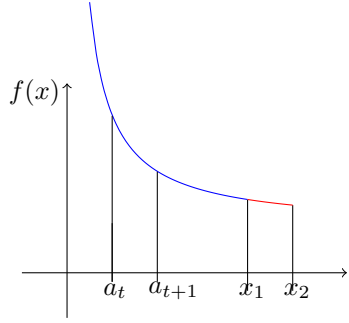
\begin{figure}[t!]
        \centering
        \begin{tikzpicture}[scale=0.6]
        \draw[-] (3.6,-0.4) node[right] {$x_1$};
        \draw[-] (4.6,-0.4) node[right] {$x_2$};
        \draw[-] (-1.5,4) node[right] {$f(x)$};
        \draw (4,-0.4) -- (4, 1.62);
        \draw (5,-0.4) -- (5, 1.5);
  \draw[->] (-1, 0) -- (6.2, 0) node[right] {};
  \draw[->] (0, -1) -- (0, 4.2) node[above] {};
          \draw[scale=0.5, domain=1:8, smooth, variable=\x, blue] plot ({\x}, { 1/\x)*10+2});  
          \draw[scale=0.5, domain=8:10, smooth, variable=\x, red] plot ({\x}, { 1/\x)*10+2});  

    \draw[-] (.6,-0.4) node[right] {$a_t$};
          \draw[-] (1.6,-0.4) node[right] {$a_{t+1}$};
    \draw (1,-0.2) -- (1, 1.1);
    \draw (2,-0.2) -- (2, 2.24);
\draw (1,-0.2) -- (1, 3.48);

 
\end{tikzpicture}
        \caption{$\chi_2 = (-\infty, x_2)$}
    \label{fig:intuition2}
\end{figure}

\begin{lemma}\label{lem:case2}
When both $f_t, g_t$ are strongly convex and smooth for all $t\le T$ with information structure as defined, with Algorithm \ref{alg}, for the case when $g_t(a_t) = 0$
$$||x_t^\star-a_{t+1}|| < c_2 ||x_t^\star-a_{t}||.$$ 
\end{lemma}
\begin{proof}
When  $g_t(a_t) = 0$, $a_t$ is on the boundary, and the chosen set is $I_t = \chi_t(a_t)$ which by definition is feasible. 
Thus,   the analysis is identical to that of Lemma \ref{lem:case1}, and we get the same relation as in Lemma \ref{lem:case1} as required.
\end{proof}

Next, we consider the final case when $g(a_t) > 0$, which is the most involved of the lot.
\begin{lemma}\label{lem:case3}
When both $f_t, g_t$ are strongly convex and smooth for all $t\le T$, $1\le t\le T, \sup_{ x\in \chi}\nabla f_t(x) \le G$ and $\sup_{ x\in \chi}\nabla g_t(x) \le G$,  with information structure as defined, with Algorithm \ref{alg}, for the case when $g_t(a_t) > 0$
$$||x_t^\star-a_{t+1}|| < c_6 ||x_t^\star-a_{t}||,$$
where $c_6 = \max\{c_2, c_3, c_4, c_5\} < 1$  for $c_4 = (1-\alpha \nu_g/L_g)^{1/2}, c_5 =  \left(1-\alpha \frac{\nu_g}{\max\{G/\textsf{dist}, L_g\}}\right)^{1/2}$.

\end{lemma}
For proving Lemma \ref{lem:case3}, we will use the strong convexity of $g_t$ as well as $f_t$.

\begin{proof}

Case a) $\delta_t = ||\frac{g_t(a_t)}{2\cL_g}|| \ge ||\nabla g_t(a_t) \frac{1}{L_g}||$, in which case the update is
\begin{equation}\label{updateg}
a_{t+1} = a_t + \alpha({\hat a}_{t} - a_t),
\end{equation}
where 
$${\hat a}_{t} = a_t - \nabla g_t(a_t) \frac{1}{L_g}.$$

Since $\delta_t = ||\frac{g_t(a_t)}{2\cL_g}|| \ge ||\nabla g_t(a_t) \frac{1}{L_g}||$, the Lipschitz condition on $g_t$ implies  that $\cB(a_t, \delta_t) \cap \chi_t = \emptyset$. Thus, $g_t({\hat a}_{t}) > 0$, i.e. ${\hat a}_{t}$ is still infeasible for $g_t$, and we want to show that  

\begin{equation}\label{dec3}
||x_t^{\star} - a_{t+1}||^2  \le c ||x_t^{\star} - a_{t}||^2,
\end{equation}
for some fixed constant $c < 1$ that does not depend on $t$. Towards this end, we will exploit the strong convexity of $g_t$.

We will connect the update \eqref{updateg} with an update Subroutine 
\textsc{Optimize} will make on a suitable initial point $x_t$, function $h$, step size $\mu$, and a feasible set $I$.
Recall that $x_t^\star = \arg \min_{x\in \chi_t} f_t(x)$.
Consider a new set $I_t' = \text{convex hull}(a_t, {\hat a}_{t}, x_t^\star)$, where $a_t$ and ${\hat a}_{t}$ are as defined in \eqref{updateg}. As discussed above, both $g_t(a_{t}) > 0$ and $g_t({\hat a}_{t}) > 0$. 
Important to note that $x_t^\star$ is not necessarily equal to $\arg \min_{x\in I_t'} g_t(x)$. However, $g_t(x_t^\star) < g_t({\hat a}_{t}) < g_t(a_{t})$ since $g_t(x_t^\star)\le 0$, while $g_t(a_{t}) > 0$ and $g_t({\hat a}_{t}) > 0$, and $g_t({\hat a}_{t}) < g_t(a_{t})$ since $\nabla g_t(a_t)$ is a descent direction for $g_t$.

Consider the update $x_{t+1}$ which Subroutine 
\textsc{Optimize} will make if the initial/starting point  $x_t=a_t$, the set $I= I_t'$ with step size $\mu=L_g$ and $h=g_t$. 
Since  ${\hat a}_{t} \in I_t'$,  ${\hat a}_{t} = \text{Proj}({\hat a}_{t}, I_t')$. Hence from Subroutine 
\textsc{Optimize} we get that 
\begin{equation}\label{updateg1}
{\hat x}_{t} = \text{Proj}(a_t - \nabla g_t(a_t) \frac{1}{L_g}, I_t') = a_t - \nabla g_t(a_t) \frac{1}{L_g} = {\hat a}_t,
\end{equation} 
\begin{equation}\label{}
x_{t+1} = a_t + \alpha({\hat x}_{t} - a_t),
\end{equation}
coinciding with \eqref{updateg}. Thus, the update of the algorithm \eqref{updateg} is equivalent to executing Subroutine 
\textsc{Optimize}  with starting point  $x_t=a_t$,  set $I= I_t'$ with step size $\mu=L_g$, for function  $h=g_t$. So we would like to use Lemma \ref{lem:mother}.
 However, since $x_t^\star$ need not be $ \arg \min_{x\in I_t'} g_t(x)$, we cannot use Lemma \ref{lem:mother} directly. Instead we exploit the fact that $g_t(x_t^\star) < g_t({\hat a}_{t}) < g_t(a_{t})$. Hence  Corollary \ref{lem:mothercor} becomes applicable, and we get 
that 
\begin{equation}\label{dummy1}
||a_{t+1} - x_t^\star||\le c_4 ||a_{t} - x_t^\star||,
\end{equation}
with $c_4 = (1-\alpha \nu_g/L_g)^{1/2}$ since we have chosen $\mu = L_g$, inverse of the step size in Subroutine 
\textsc{Optimize}.




Case b) $\delta_t = ||\frac{g_t(a_t)}{2L_g}|| < ||\nabla g_t(a_t) \frac{1}{L_g}||$ 

In this case, we have no sufficiently sized estimate of 
the infeasible region around $a_t$. Thus, we will exploit the strong convexity and smoothness of $f_t$, as follows. 

b-i) Let $\chi_t(a_t) \neq \emptyset$.
In this case, $I_t= \chi_t(a_t)$,
		$a_t' = \text{Proj}(a_t, I_t)$ and $a_{t+1}=\textsc{Optimize}(f_t, I_t,\frac{1}{2L_f},a_t')$. 
		
		Subcase b-i-i)  Let $I_t= \chi_t(a_t) \subseteq \text{int}(\cB(a_t, \textsf{dist}))$ which implies that 
$x_t^\star \in I_t$. Recall that by definition, $a_t' = \text{Proj}(a_{t}, I_t)$. Thus, 
with $x_t^\star \in I_t$, we get directly from Lemma \ref{lem:mother} that 
$$||a_{t+1} - x_t^\star || \le c_2 ||a_t' - x_t^\star|| \le  c_2 ||a_t - x_t^\star||,$$
where the last inequality follows since $a_t \notin I_t$.

Subcase b-i-ii) Let $I_t= \chi_t(a_t) \not\subset \cB(a_t, \textsf{dist})$. In this sub-case, we get that $g_t(a_t') =0$ and is identical to the  case considered in Lemma \ref{lem:case2}, except the starting point is $a_t'$ instead of $a_t$. Thus, similar to \eqref{dummy1xx}, we get the first inequality 
\begin{equation}\label{dummy3}
||x_t^{\star} - a_{t+1}|| \le \max\{c_2,c_3\} ||x_t^{\star} - a_{t}'|| \le  \max\{c_2,c_3\}  ||x_t^{\star} - a_{t}||,
\end{equation} 
where the second inequality follows since $a_t \notin I_t$.

Case b-ii) Let $\chi_t(a_t) = \emptyset$. In this case, the update is 
 \begin{equation}\label{infeaslast}
a_{t+1} =   a_t + \alpha ({\hat a}_{t} - a_t),
\end{equation} where  
${\hat a}_{t} = a_t- \nabla g_t(a_t)~\frac{\textsf{dist}}{||\nabla g_t(a_t)||}$. Since $\chi_t(a_t) = \cB(a_t, \textsf{dist}) \cap \chi_t$ is empty, $g_t( {\hat a}_{t}) > 0$. Thus, we can 
exploit the strong convexity and smoothness of $g_t$ as in case a). 


Given the assumption that $\nabla g_t(a_t) \le G$, we get that 
\eqref{infeaslast} is equivalent to executing \textsc{Optimize} with set $I_t = \text{convex hull}(a_t , {\hat a}_{t}, x_t^\star)$, $h=g_t$, and $\mu = \max\{G/\textsf{dist}, L_g\}$. Thus, similar to \eqref{dummy1}, since $g_t( x_{t}^\star) < g_t( {\hat a}_{t}) < g_t(a_{t})$, we get 
\begin{equation}\label{infeasfinrelation}
||a_{t+1} - x_t^\star||\le c_5 ||a_{t} - x_t^\star||,
\end{equation}
where $c_5 =  \left(1-\alpha \frac{\nu_g}{\max\{G/\textsf{dist}, L_g\}}\right)^{1/2}$.

\section{Conclusions}
In this paper, we considered a constrained OCO problem, and provided the best (simultaneously) possible bounds for the regret and the constraint violation penalty, when both the loss function and the function defining the constraint are strongly convex and smooth. Compared to prior work, we proposed an algorithm that has better regret and penalty bounds while using significantly less information requirement about the loss function and the function defining the constraints. Extending these results when the respective functions are just convex and not strongly convex, remains open.

\end{proof}

\bibliographystyle{IEEEtran}
\bibliography{../refs}
\section{Proof of Corollary \ref{lem:mothercor}}\label{appcor}
Using the $\nu_h$-strong convexity, and $L$ smoothness of function $h$, from (31) \cite{cdcstrictconvex}, we have that for any $x \in I$ and $x_t \in I$ as the starting point and ${\hat x}_t$ as defined in \textsc{Optimize}, for $\mu \ge L$,  $h(x) - \frac{\nu_h}{2 } || x- x_t||^2$
\begin{equation}\nn
 \ge h({\hat x}_t) + \frac{\mu}{2 } || {\hat x}_t- x_t||^2 + \mu (x_t - {\hat x}_t)^T(x  - x_t).
\end{equation}
Choosing $x= x' \in I$ such that $h(x') < h( {\hat x}_t)$, and rearranging terms, we get that $h(x') - h({\hat x}_t)$
\begin{equation}\nn
  \ge   \frac{\mu}{2 } || {\hat x}_t- x_t||^2  + \frac{\nu_h}{2 } || x'- x_t||^2 + \mu (x_t - {\hat x}_t)^T(x' -  x_t).
\end{equation}
Using the fact that $h(x) < h( {\hat x}_t) < h(x_t)$, the LHS  is negative, and dividing both sides by $\mu$, and rearranging terms we get  
\begin{equation}\label{dummyy1}
(x_t - {\hat x}_t)^T(x_t - x') \ge \frac{1}{2}|| {\hat x}_t- x_t||^2 + \frac{\nu_h}{2 \mu}|| x'- x_t||^2.
\end{equation}

Recall that $x_{t+1} = (1-\alpha)x_t + \alpha {\hat x}_t$. Using this, $||x_{t+1} - x'||^2 $
\begin{equation}\nn
= ||x_t - x'||^2 +\alpha^2 ||x_t - {\hat a}_{t}||^2 - 2\alpha (x_t - x')^T(x_t -  {\hat x}_t).
\end{equation}
Using the bound on $ (x_t - x')^T(x_t -  {\hat x}_t)$ obtained in \eqref{dummyy1}, we get 
\begin{equation}\nn
||x_{t+1} - x'||^2 \le \left(1-\frac{\alpha \nu_h}{\mu}\right) ||x_t - x'||^2 +\alpha(\alpha-1) ||x_t - {\hat a}_{t}||^2.
\end{equation}
Since $\alpha \in (0,1]$, the second term in RHS is non-positive. Thus, we get \begin{equation}\label{}
||x_{t+1} - x'||^2 \le \left(1-\frac{\alpha \nu_h}{\mu}\right) ||x_t - x'||^2,
\end{equation}
as required.
 	\end{document}